\documentclass[letterpaper, 10 pt, conference]{ieeeconf}  
\IEEEoverridecommandlockouts   
\usepackage{times}
\usepackage{epsfig}
\usepackage{graphicx}
\usepackage{amssymb}
\usepackage{amsmath}
\usepackage{amsthm}

\usepackage[dvipsnames]{xcolor}

\usepackage{gensymb}
\usepackage[hyphens]{url}
\usepackage{multirow}
\usepackage{multicol}
\usepackage{tabularx}
\usepackage{dsfont}
\usepackage{url}
\usepackage{color}
\usepackage{subcaption}
\usepackage{caption}
\usepackage{booktabs}
\usepackage{dcolumn}
\usepackage{colortbl}
\usepackage{pifont}
\usepackage{algorithm}
\usepackage{xspace}
\usepackage{algpseudocode}

\usepackage[pagebackref=true,breaklinks=true,letterpaper=true,colorlinks,bookmarks=false]{hyperref}

\colorlet{lightgray}{gray!20}

\makeatletter
\DeclareRobustCommand\onedot{\futurelet\@let@token\@onedot}
\def\@onedot{\ifx\@let@token.\else.\null\fi\xspace}

\def\eg{\emph{e.g}\onedot} 
\def\ie{\emph{i.e}\onedot}

\begin{document}

\title{V2X-ReaLO: An Open Online Framework and Dataset for Cooperative Perception in Reality}

\author{Hao Xiang, Zhaoliang Zheng, Xin Xia, Seth Z. Zhao, Letian Gao, Zewei Zhou, Tianhui Cai, Yun Zhang, Jiaqi Ma\\
University of California, Los Angeles 
}

\maketitle

\begin{abstract}
Cooperative perception enabled by Vehicle-to-Everything (V2X) communication holds significant promise for enhancing the perception capabilities of autonomous vehicles, allowing them to overcome occlusions and extend their field of view. However, existing research predominantly relies on simulated environments or static datasets, leaving the feasibility and effectiveness of V2X cooperative perception especially for intermediate fusion in real-world scenarios largely unexplored. In this work, we introduce V2X-ReaLO, an open online cooperative perception framework deployed on real vehicles and smart infrastructure that integrates early, late, and intermediate fusion methods within a unified pipeline and provides the first practical demonstration of online intermediate fusion's feasibility and performance under genuine real-world conditions. Additionally, we present an open benchmark dataset specifically designed to assess the performance of online cooperative perception systems. This new dataset extends V2X-Real dataset to dynamic, synchronized ROS bags and provides 25,028 test frames with 6,850 annotated key frames in challenging urban scenarios. By enabling real-time assessments of perception accuracy and communication lantency under dynamic conditions, V2X-ReaLO sets a new benchmark for advancing and optimizing cooperative perception systems in real-world applications. The codes and datasets will be released to further advance the field.

 \end{abstract}

\maketitle
\section{Introduction}\label{sec1}
As autonomous driving continues to progress, ensuring robust perception in complex traffic environments remains a critical objective. Recent advances in deep learning have significantly improved single-vehicle perception~\cite{mei2022waymo, paz2020probabilistic, yurtsever2020survey, yang2023bevformer, hu2023planning, grigorescu2020survey, caesar2020nuscenes} for tasks such as object detection~\cite{lang2019pointpillars, carion2020end, yang2018pixor, li2022bevformer, liu2023bevfusion} and tracking~\cite{liang2020pnpnet, li2021visio}, yet these systems remain constrained by limited viewpoints, sparse sensor observations, and occlusions, ultimately restricting their situational awareness. To address these issues, Vehicle-to-Everything (V2X) cooperative perception~\cite{xu2022opv2v, xu2022v2x, xiang2023hm, xu2022cobevt, wang2020v2vnet} has emerged as a promising approach, enabling vehicles and infrastructure to exchange diverse sensory information (\eg, raw LiDAR point clouds, detection outputs, or neural features) for a more holistic representation of the environment. 

\begin{figure}
    \centering
    \includegraphics[width=1\linewidth]{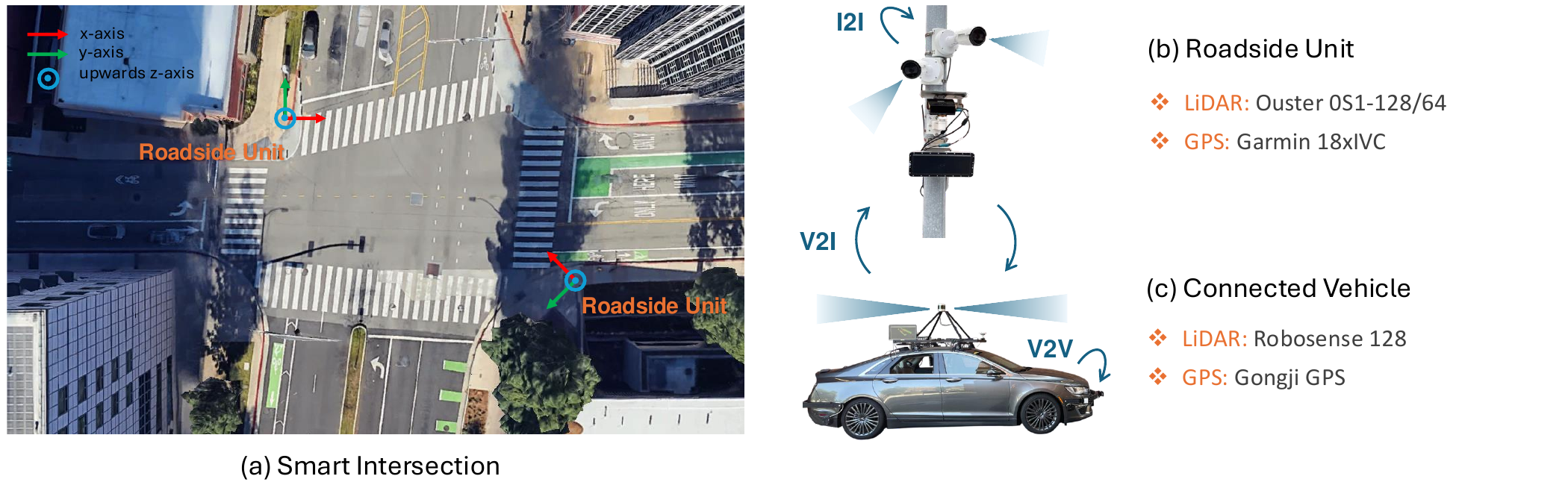}
    \caption{\textbf{Online testing system for real-time cooperative perception.} (a) Smart intersection. (b) Roadside unit equipped with Ouster OS1-128/64 LiDAR and Garmin GPS. (c) Connected vehicle retrofitted with a 128-channel Robosense LiDAR and Gongji GPS.  Wireless V2X communication employs Wi-Fi network..}
    \label{fig:testbeds}
\end{figure}

Despite its great potential, existing research~\cite{wang2020v2vnet, xu2022opv2v, chen2019f, zhao2024coopre, xu2023v2v4real, xiang2024v2x, zhou2024v2xpnp} primarily focus on offline evaluations based on simulation dataset~\cite{li2022v2x, xu2022opv2v, xu2022v2x} or static real-world datasets~\cite{xiang2025v2x, xu2023v2v4real, yu2022dair, hao2024rcooper, yu2023v2x}. While these studies offer valuable insights for algorithm development,  they fall short of addressing the complexities inherent in real-time deployment where challenges such as transmission latency, limited computation, localization errors can compound to significantly impact the performance of V2X systems~\cite{xu2022v2x, xu2021opencda, vadivelu2021learning}. For example,  pose estimation errors can cause spatiotemporal misalignments among the fused feature maps and these misalignments are further exacerbated by latency spikes, ultimately resulting in incomplete or inconsistent scene representations that undermines the effectiveness of cooperative perception.
Although recent works have established standards~\cite{SAEJ3216, SAEJ3295} and practical examples~\cite{rauch2011analysis, shan2020demonstrations, kim2014multivehicle, zheng2024cooperfuse, song2024first} for transmitting detection outputs (Late Fusion) and raw sensor data (Early Fusion), it remains unclear \emph{whether and how} intermediate features can be exchanged in real-world scenarios to improve perception performance. Implementing a robust online V2X intermediate fusion pipeline involves several novel challenges: neural features reside in GPU memory and must be transmitted in real time, necessitating robust host–device data transfers, serialization, transmission standards, and compression. Yet existing studies seldom investigate the feasibility of these high-dimensional tensor exchanges under real-world bandwidth and latency constraint    s, nor do they propose methodologies or protocols for managing such abstract neural features. Consequently, it remains unknown whether intermediate fusion can truly enhance perception systems in real-world deployments, thus limiting its scalability for large-scale autonomous driving.

Moreover, conducting real-world experiments in online cooperative perception demands substantial logistical and technical resources. Deploying such systems necessitates the integration of sophisticated sensor suites, high-performance computing platforms, and robust V2X communication. Furthermore, coordinating and executing multi-vehicle experiments in dynamic real-world environments requires considerable human and logistical resources. These substantial resource requirements limit the ability of researchers to thoroughly investigate and validate their algorithms in realistic scenarios, thus impeding the progress of online cooperative perception research.



In this work,  we introduce V2X-ReaLO, an open online cooperative perception framework that can be deployed on real vehicles and infrastructure. 
 Built on the Robot Operating System (ROS)~\cite{quigley2009ros}, V2X-ReaLO integrates early, late, and intermediate fusion strategies within a unified pipeline. Crucially, V2X-ReaLO provides the first concrete demonstration of the feasibility, effectiveness, and performance of intermediate fusion under genuine real-world constraints. 
In addition, we present an open online benchmark dataset specifically designed for online evaluations. This new dataset extends the test split of V2X-Real~\cite{xiang2024v2x} dataset, to dynamic, synchronized ROS bags with 25,028 frames, of which 6,850 are fully annotated key frames in challenging urban scenarios. By leveraging this dataset and the proposed online framework, researchers can more readily develop and evaluate novel online cooperative perception algorithms without the need for expensive hardware deployments, thereby lowering the barrier to entry for online experimentation and facilitating rapid prototyping, development, and comparative analysis.  Finally, we further conduct comprehensive experiments to evaluate the system’s online performance under real-world bandwidth and latency constraints. All associated codes and datasets will be publicly released, fostering continued research and innovation in the field. Our contributions can be summarized:

\begin{itemize}
    \item We introduce V2X-ReaLO, an open {online} cooperative perception framework deployable in real vehicles and infrastructure. It integrates early, late and intermediate fusion methods in a unified framework and provides the first practical demonstration of intermediate fusion’s feasibility and performance under real-world conditions. 
    \item We extend V2X-Real to dynamic, synchronized ROS bags, featuring 25,028 frames, of which 6,850 are fully annotated key frames. This dataset enables real-time evaluations of bandwidth, latency, and perception accuracy in complex urban scenarios without requiring substantial hardware deployments.
    \item We conduct extensive benchmarks for {online} multi-class, multi-agent V2X cooperative perception under various collaboration modes (\ie, V2V, V2I, and I2I), illustrating the system’s effectiveness in real-world settings. 
\end{itemize}


\begin{figure*}
    \centering
    \includegraphics[width=1\linewidth]{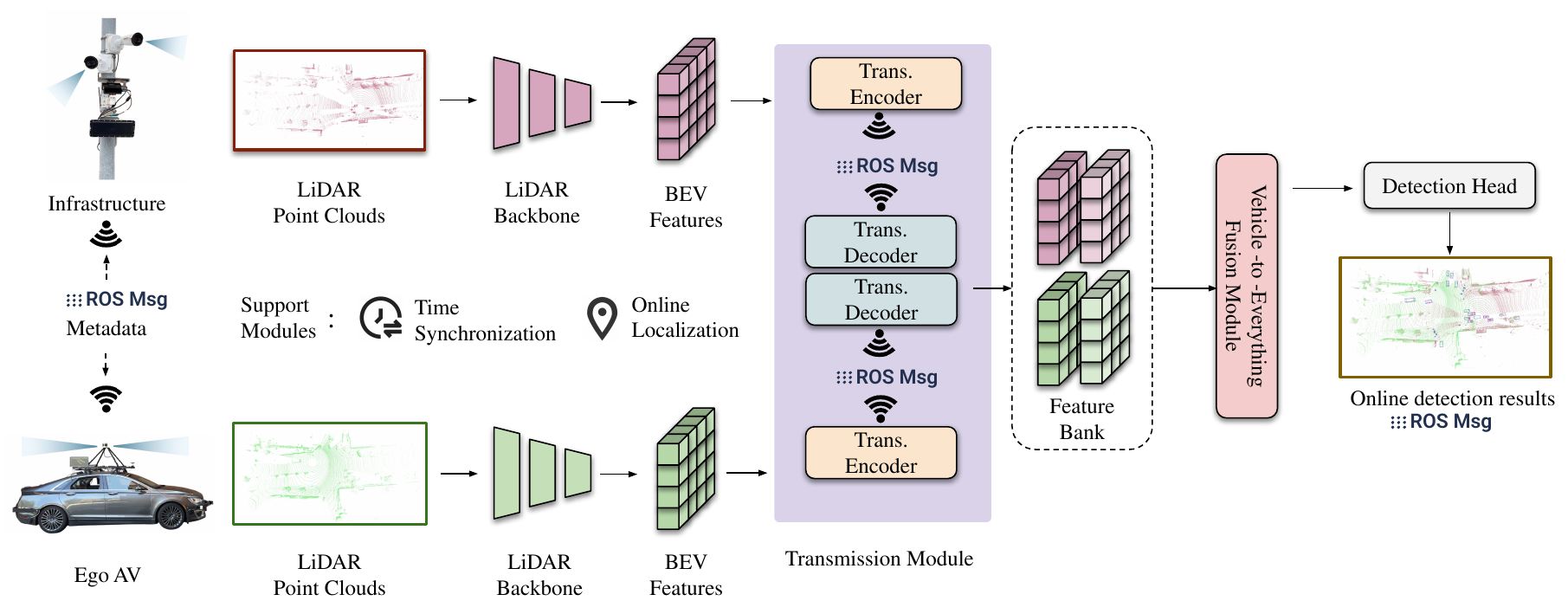}
    \caption{\textbf{Framework of online intermediate fusion.} Each agent processes LiDAR point clouds to generate Bird’s Eye View (BEV) features, which are compressed by a Transmission Encoder and transmitted over the Wi-Fi network as ROS messages. On the receiving side, these features are reconstructed via Transmission Decoder, stored in a Feature Bank, and retrieved by the ego agent for fusion with its own BEV representation. Finally, the fused features are passed to a Detection Head for perception outputs. The entire pipeline operates in ROS, with additional support modules (\eg, time synchronization, online localization) ensuring seamless multi-agent coordination.}
    \label{fig:framework}
\end{figure*}

\section{Related work}
\subsection{Cooperative Perception}
Cooperative perception seeks to enhance an agent’s situational awareness by sharing and integrating sensing information with other connected agents. Based on the type of shared data, existing cooperative perception algorithms can be broadly categorized into three paradigms: late fusion, in which detection outputs are exchanged~\cite{rawashdeh2018collaborative}; early fusion, which circulates raw LiDAR point clouds~\cite{chen2019cooper}; and intermediate fusion, where neural features are transmitted among agents~\cite{xiang2023hm, xu2022v2x, xu2022cobevt}. Early research primarily investigated late and early fusion. While late fusion minimizes bandwidth usage, it can result in information loss during transmission, leading to compounded errors in fusion. 
 In contrast, early fusion preserves all raw data but incurs substantial bandwidth demands, making it impractical for large-scale, real-time deployment.  Recent studies have increasingly focused on intermediate fusion, which balances accuracy and bandwidth efficiency by transmitting compressed feature representations rather than raw data or final detections. Following this line of research, OPV2V~\cite{xu2022opv2v} presented an attentive intermediate fusion network for vehicle-to-vehicle cooperative perception and V2X-ViT~\cite{xu2022v2x} further advanced the field by introducing a vision transformer architecture tailored for Vehicle-to-Everything (V2X) cooperative perception. DiscoNet~\cite{li2021learning} demonstrated a distillation method for learning multi-agent collaborations. HM-ViT~\cite{xiang2023hm} proposed a heterogeneous vision transformer for multi-agent multi-modal sensor fusion. Despite the promising performance of these Intermediate Fusion approaches, most evaluations remain confined to simulation or offline settings, where challenges such as transmission latency, limited computational resources, and asynchronous message arrivals are not fully captured.  Consequently, it remains uncertain whether and how intermediate fusion can operate robustly in real-world scenarios,  thereby hindering its deployment.


\begin{figure}
    \centering
    \includegraphics[width=0.8\linewidth]{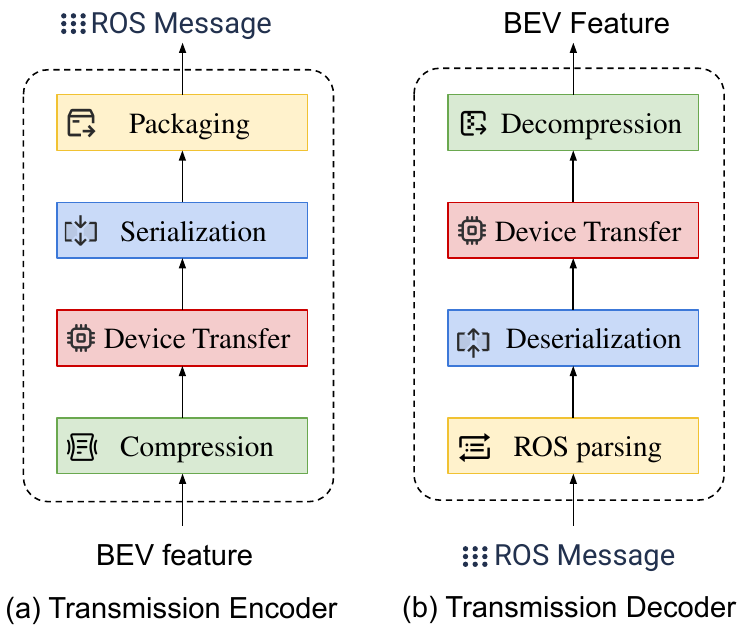}
    \caption{\textbf{Pipeline of transmission modules.} (a) Transmission Encoder compresses the BEV feature, transfers it from the GPU to the host, serializes the data, and packages it into a ROS message. (b) Transmission Decoder performs the reverse operations by parsing the ROS message, deserializing and transferring the data back to the GPU, and finally decompressing the feature.}
    \label{fig:transmission_module}
\end{figure}

\subsection{Cooperative Perception Dataset}
Recent advancements in cooperative perception datasets have significantly contributed to the field of multi-agent perception. OPV2V~\cite{xu2022opv2v} pioneered simulation datasets for Vehicle-to-Vehicle (V2V) cooperative perception, leveraging the CARLA simulator~\cite{dosovitskiy2017carla} to support algorithm development in controlled environments. V2X-Set~\cite{xu2022v2x} and V2X-Sim~\cite{li2022v2x} expanded these efforts by integrating additional infrastructure sensor data, facilitating more complex Vehicle-to-Infrastructure (V2I) interactions in simulated environment. These datasets facilitated the initial development and evaluation of cooperative perception algorithms within controlled, synthetic environments. The field has since shifted toward real-world datasets. DAIR-V2X~\cite{yu2022dair} introduced a real-world Vehicle-to-Infrastructure dataset, while V2V4Real~\cite{xu2023v2v4real} provided a real-world dataset focused on V2V cooperative perception. V2X-Real~\cite{xiang2024v2x} further advances this field by offering a comprehensive, large-scale dataset with full collaboration modes (\ie, V2V, V2I, I2I) for V2X cooperative perception, featuring two connected vehicles and two roadside units. However, existing datasets are designed for offline evaluations, assuming ideal or constant conditions, and thus do not effectively capture the real-time challenges such as bandwidth limitations, latency, and asynchronous message arrival, that are crucial for the online performance of cooperative perception. In contrast, our dataset focuses on online evaluation by extending V2X-Real to ROS bags, thus enabling a more realistic assessment of how cooperative perception operates under dynamic, real-world constraints.


\section{Online Intermediate Fusion Framework}

 Our framework supports all three fusion strategies, \ie, early, late, and intermediate fusion. In this section, we detail the architecture of our online intermediate fusion; the pipelines for late and early fusion are described in Sec.~\ref{sec:benchmarked_methods} and detailed in supplementary material. The entire framework is implemented using the Robot Operating System (ROS)~\cite{quigley2009ros}, a standard middleware that facilitates distributed node communication, sensor data integration, and real-time replay. Fig.~\ref{fig:framework} presents an overview of the proposed online intermediate fusion framework, wherein each agent operates as a ROS node. Each node acquires local sensor observations (e.g., LiDAR scans, agent poses), exchanges intermediate representations with collaborators' ROS nodes over the network, and ultimately produces final detection outputs through a fusion module.

\subsection{Main Architecture}

\noindent\textbf{Synchronization:} Prior to real-time collaboration, the time clocks of all agents' host machines are synchronized with GPS time, and the LiDAR sensors are phase-locked with GPS signals. This ensures temporal synchrony across all agents' sensing observations, which is essential for accurate data integration and collaboration.

\noindent\textbf{Localization: }To achieve robust positioning for each agent, we first construct a dense LiDAR point cloud map of the full driving route, following~\cite{xia2023automated}. During online testing, we perform map matching in conjunction with GPS/IMU signals~\cite{gao2023gnss}, aligning each agent’s LiDAR scans to the pre-built map. This process yields an accurate transformation between the agent’s local LiDAR frame and the global map coordinate system, ensuring a consistent spatial reference for subsequent cooperative perception tasks.

\noindent\textbf{Metadata Sharing:} During the initial stage of collaboration, each agent continuously shares metadata, including poses and timestamps, with other agents via ROS messages which will be stored in a buffer on the receiving end. Upon receiving the LiDAR point cloud message,  each agent then retrieves the most recent ego agent's pose information from the buffer and uses it to project its LiDAR point clouds into the ego agent's coordinate frame prior to feature extraction.

\begin{table}[ht]
    \centering
    \begin{tabular}{l|l}
        \toprule
        {Method} & {ROS Message Fields} \\ \hline
        Late Fusion &
        \begin{tabular}[t]{@{}l@{}}
            \texttt{MarkerArray boxes}
        \end{tabular} \\ \hline
        Early Fusion &
        \begin{tabular}[t]{@{}l@{}}
            \texttt{PointCloud2 points}
        \end{tabular} \\ \hline
        Intermediate Fusion &
        \begin{tabular}[t]{@{}l@{}}
            \texttt{Header header} \\
            \texttt{UInt8MultiArray bev}
        \end{tabular} \\ \bottomrule
    \end{tabular}
    \caption{\textbf{ROS messages for late, early, and intermediate fusion strategies.}}
    \label{tab:ros_messages}
\end{table}

\noindent\textbf{Feature Extraction:} We adopt  {PointPillar}~\cite{lang2019pointpillars} to process each agent's point clouds by voxelizing it into pillars along the $x$ and $y$ axes. A lightweight neural network encodes each pillar’s geometric information, producing a BEV feature map $F \in \mathbb{R}^{H \times W \times C}$, which is then passed to the transmission encoder for network communication.  

\noindent\textbf{Transmission Module:} After obtaining BEV features, the transmission encoder compresses and serializes these high-dimensional feature maps into ROS messages for network transfer. On the receiving side, the transmission decoder parses and deserializes the incoming ROS messages to retrieve the compressed features, then decompresses them back to their original dimensionality. The recovered features are subsequently stored in a memory buffer dubbed \emph{Feature Bank}. The details of the transmission module are provided in Sec.~\ref{sec:transmission_module}.

\noindent\textbf{V2X Fusion: }When ego agent finishes extracting the BEV features from its own sensing observations, it queries the  {Feature bank} for the latest collaborator's neural messages that fall within a specific time range relative to the current timestamp. This ensures that the collaborator's messages are temporally close and therefore informative. The queried collaborator's intermediate features and ego agent's BEV features are then passed to a fusion module for refinement. 

\noindent\textbf{Detection Head: }The final refined features are passed to a multi-class head to produce classification scores and bounding box proposals. During training, we adopt a smooth $\ell_1$ loss for bounding box regression and a focal loss for classification. At inference, we apply Non-Maximum Suppression (NMS) to fuse overlapping proposals into unified predictions.  Notably, our dataset can feature over 100 active road users in a single scene, causing the computational burden for NMS (whose pairwise IoU calculations scale quadratically with the number of boxes) to rise sharply. To alleviate this overhead, we adopt an R-tree data structure that organizes bounding boxes by spatial proximity, restricting IoU checks to smaller subsets of nearby objects.  This approach substantially reduces computation while preserving real-time performance in dense urban scenarios.

\subsection{Transmission Module}
\label{sec:transmission_module}
In an online multi-agent setting, efficient data exchange among collaborators is essential for real-time cooperative perception, yet existing studies rarely investigate this aspect. To address this gap, we introduce a  {Transmission Module} that manages feature compression, data transfer, serialization, and network communication, maintaining feasible bandwidth usage while preserving the fidelity required for robust perception.  As illustrated in Fig.~\ref{fig:transmission_module}, this module consists of two core components: a  {transmission encoder} at the sending agent and a  {transmission decoder} at the receiving agent. The encoder reduces the channel dimensionality of Bird’s Eye View (BEV) features and serializes them for network transfer, whereas the decoder reconstructs these features on the receiving side, restoring them to their original dimensionality and placement in GPU memory.

\noindent\textbf{Transmission Encoder.}
After obtaining the BEV features from the LiDAR backbone, we apply a lightweight compression network composed of consecutive 3×3 convolutions, batch normalization, and ReLU activations to reduce the channel dimension by a factor of compression ratio. The compressed features, residing in GPU memory, are then transferred to the host (CPU) and serialized into a one-dimensional byte stream. Subsequently, this serialized data, along with relevant metadata such as headers and agent identifiers, is encapsulated in a ROS message (as shown in Tab.~\ref{tab:ros_messages}) and dispatched to other connected agents within a defined spatial range of the collaboration graph.

\noindent\textbf{Transmission Decoder.}
On the receiving end, the  {Transmission Decoder} parses and deserializes the incoming ROS messages, retrieving both the compressed feature and any associated metadata.The data are then transferred back to GPU memory and fed into a decompression network featuring another series of 3×3 convolutions, batch normalization, and ReLU layers, restoring the original channel dimensionality. These reconstructed features, along with metadata, are stored in a  {Feature Bank}, ensuring the ego agent can retrieve features that are temporally close.

\noindent\textbf{Communication: }In this study, we adopt a Wi-Fi network for multi-agent communication and employ Swarm ROS bridge~~\cite{swarm} for transmitting customized ROS messages. Notably, our framework is readily extendable to other communication infrastructures that support ROS protocols.

\begin{table*}[!t]
\centering
\begin{tabular}{@{}ccc@{}}
\begin{minipage}[t]{0.33\textwidth}
\begin{tabular}{c|c|c}
         \toprule
         Comp. Rate&Msg. size&Latency \\
         \midrule
         0x&25MB&511.2ms\\
         8x&3.1MB&97.0 ms \\
         32x&840KB&41.7ms\\
         64x&391.1KB&14.3ms\\
         \bottomrule
    \end{tabular}
    \caption{\textbf{Compression rate ablation study.} }
    \label{tab:compression_rate}
\end{minipage}%
&
\begin{minipage}[t]{0.37\textwidth}
\begin{tabular}{c|c|c}
    \toprule
         Modules&Encoder&Decoder  \\
        \midrule
         Compression&0.13ms&0.27ms\\
         Device Transfer&19.19ms&0.43ms\\
         Serialization&0.26ms&0.07ms\\
         Packaging/Parsing&0.01ms&0.01ms\\
        \bottomrule
    \end{tabular}
    \caption{\textbf{Computation time for transmission encoder and decoder.}}
    \label{tab:transmission_latency}
\end{minipage}%
&
\begin{minipage}[t]{0.26\textwidth}
\begin{tabular}{c|c}
         \toprule
         Models&Inf. Speed  \\
         \midrule
         Late Fusion&43.7ms\\
         Early Fusion&80.9ms \\
         Fcooper~\cite{chen2019cooper}&61.6ms\\
         AttFuse~\cite{xu2022opv2v}&68.6ms\\
         V2X-ViT~\cite{xu2022v2x}&118.2ms\\
         \bottomrule
    \end{tabular}
    \caption{\textbf{Inference speed study. }}
    \label{tab:inference_speed}
\end{minipage}%
\end{tabular}
\end{table*}

\begin{table}[]
    \centering
    \begin{tabular}{c|c|c}
         \toprule
         Message type&Message Size&Latency  \\
         \midrule
          Detection outputs&3.63KB&10.0ms\\
         LiDAR points&3.7MB&144.2ms \\
         Intermediate Features (32x)&840KB&41.7ms\\
         \bottomrule
    \end{tabular}
    \caption{\textbf{Message size and transmission latency experiments.} }
    \label{tab:latency}
\end{table}



\section{Online Cooperative Perception dataset}


Our {online} benchmark dataset is built upon the V2X-Real dataset~\cite{xiang2024v2x}, a large-scale, real-world dataset designed for V2X cooperative perception. V2X-Real was collected from two connected automated vehicles (CAVs) and two roadside units (RSUs) operating in various urban scenarios (\eg, V2X intersections, V2V corridors). During data collection, raw sensor outputs for each agent, including LiDAR scans, stereo camera images, and GPS signals, were recorded in ROS bags, the Robot Operating System (ROS)~\cite{quigley2009ros} 
logging format that supports time-stamped storage and replay of sensor data.  During post-processing, LiDAR frames and camera images were extracted from these raw ROS bags, resulting in over 33{,}000 LiDAR frames and 171{,}000 multi-view camera images, which were annotated with more than 1.2 million 3D bounding boxes spanning 10 object categories. The V2X-Real dataset is partitioned into training (23,379 frames), validation (2,770 frames), and testing (6,850 frames) splits, providing a comprehensive basis for offline V2X cooperative perception development.



 Our online benchmark adapts the {test split} of V2X-Real to facilitate real-time testing and latency measurements for online cooperative perception. For each agent’s annotated driving log, we select the corresponding ROS bag, which contains annotated key frames and supplementary context frames. We then synchronize these agent-specific ROS bags via GPS time and merge them into a single unified bag that consolidates multi-agent sensor data. The resulting ROS bag comprises 25{,}028 frames, including 6{,}850 fully annotated key frames spanning complex urban environments. These data encapsulate diverse collaboration modes \eg, Vehicle-to-Vehicle (V2V), Vehicle-to-Infrastructure (V2I), and Infrastructure-to-Infrastructure (I2I), thus capturing a broad range of V2X interaction patterns.

By replaying sensor messages at real-time rates, our {online} dataset addresses the limitations of {offline} evaluations, which often assume static or idealized conditions. In practical deployments, factors such as asynchronous message arrivals, transmission latency, and resource contention can interleave in complex ways that static datasets cannot capture. For example, a slower perception model can cause buffering or frame skips during online operation, potentially discarding the matched collaborator messages due to limited buffer size and enlarging the time gap between the ego agent’s and collaborator’s observations. Such discrepancies further exacerbate misalignments, highlighting the importance of measuring real-time performance in cooperative perception systems. Consequently, beyond merely demonstrating feasibility, our {online} dataset and benchmarks provide a more faithful environment for examining how these real-world constraints affect perception accuracy, offering deeper insights into the trade-offs of large-scale V2X cooperative perception development—an aspect often overlooked in existing works.

\section{Online Evaluation} 
\begin{table*}[]
    \centering
    \begin{tabular}{c|c|cccccc|cc}
    \toprule
       \multirow{2}{*}{Dataset} &\multirow{2}{*}{Models}& \multicolumn{2}{c}{AP$_{car}$@IoU} & \multicolumn{2}{c}{AP$_{ped.}$@IoU} &  \multicolumn{2}{c|}{AP$_{truck}$@IoU}& \multicolumn{2}{c}{mAP@IoU}  \\
        & & 0.3 & 0.5 & 0.3 & 0.5 & 0.3 & 0.5 & 0.3 & 0.5\\
        \midrule
        \multirow{6}{*}{V2X-ReaLO-V2V}&No Fusion&41.2&37.6&\textbf{24.7}&\textbf{12.7}&21.4&13.9&29.1&21.4 \\
        \cmidrule{2-10}
        &Late Fusion&41.6&32.0&18.1&8.4&16.4&9.4&25.3&16.6\\
        \cmidrule{2-10}
        &Early Fusion&41.0&31.7&17.4&8.3&20.3&12.4&26.2&17.5\\
        \cmidrule{2-10}
        &F-Cooper~\cite{chen2019cooper}&54.3&43.0&18.0&6.7&{36.5}&27.0&36.3&25.6 \\
        &AttFuse~\cite{xu2022opv2v}&\textbf{54.7}&\textbf{45.9}&{21.7}&{9.6}&{34.5}&{26.6}&\textbf{37.0}&\textbf{27.4} \\
        &V2X-ViT~\cite{xu2022v2x}&48.2&40.3&21.4&9.1&\textbf{36.8}&\textbf{30.5}&35.5&26.6 \\
        \midrule
        \multirow{6}{*}{V2X-ReaLO-V2I}&No Fusion&41.0&37.6&{25.8}&{13.5}&21.9&15.7&29.6&22.3 \\
        \cmidrule{2-10}
        &Late Fusion&47.1&43.1&26.7&13.1&21.1&17.0&32.3&24.4\\
        \cmidrule{2-10}
        &Early Fusion&50.1&42.9&27.2&13.7&32.4&23.6&36.6&26.7\\
        \cmidrule{2-10}
        &F-Cooper~\cite{chen2019cooper}&52.9&45.0&22.2&9.1&25.9&18.3&33.7&24.1\\
        &AttFuse~\cite{xu2022opv2v}&\textbf{53.3}&\textbf{48.7}&\textbf{28.3}&\textbf{14.1}&{32.4}&{28.0}&\textbf{38.0}&\textbf{30.3} \\
        &V2X-ViT~\cite{xu2022v2x}&49.2&43.6&19.9&9.0&\textbf{41.0}&\textbf{35.6}&36.7&29.4\\
        \midrule
        \multirow{6}{*}{V2X-ReaLO-I2I}&No Fusion&48.1&39.4&30.5&15.3&23.3&22.4&34.0&25.7 \\
        \cmidrule{2-10}
        &Late Fusion&63.4&53.6&35.0&15.4&36.3&26.4&44.9&31.8\\
        \cmidrule{2-10}
        &Early Fusion&60.9&57.2&\textbf{41.2}&\textbf{21.9}&38.5&30.8&46.9&36.6 \\
        \cmidrule{2-10}
        &F-Cooper~\cite{chen2019cooper}& 56.1&49.3&33.5&12.8&\textbf{48.6}&39.9&46.0&34.0\\ 
        &AttFuse~\cite{xu2022opv2v}&{67.0}&{61.0}&{40.5}&{20.2}&{47.6}&\textbf{42.4}&{51.7}&{41.2} \\
        &V2X-ViT~\cite{xu2022v2x}&\textbf{78.0}&\textbf{70.5}&{38.8}&15.8&42.3&40.0&\textbf{53.1}&\textbf{42.1}\\
        \bottomrule
    \end{tabular}
    \caption{\textbf{Online benchmark results for V2X-ReaLO-V2V, V2X-ReaLO-V2I, V2X-ReaLO-I2I.}}
    \label{tab:benchmark}
\end{table*}

\noindent\textbf{Latency Measurements: }
 To accurately measure the network latency of transmitted V2X messages, we utilize a round-trip latency measurement approach over a wifi network. Each device first synchronizes its local clock with a global time server, ensuring consistent timeframes across all agents. ROS communication is then established, and the relevant data (\eg, LiDAR point clouds, detection outputs, or intermediate features) are circulated through the network by replaying the ROS bags. To quantify latency, we record the time required for data to complete a round-trip between two machines and then halve this duration to obtain the average one-way latency. Multiple data exchanges are performed to obtain robust, averaged statistics.

\noindent\textbf{Perception Performance Evaluation: }
We simultaneously replay the online dataset ROS bags and activate the cooperative perception ROS nodes in an online fashion, enabling the model  to consume both live sensor data and collaborator messages to produce detection outputs. These outputs, along with their timestamps, are then recorded locally. After the system has generated predictions for all timestamps, we post-process the recorded results to align them with the ground truth targets at the corresponding timestamps. Afterwords, these aligned results are used to calculate the evaluation metrics. To facilitate evaluation on a single machine, we incorporate the measured latency (as shown in Tab.~\ref{tab:latency}) into the online perception algorithm, ensuring the ego agent discards collaborator messages that arrive earlier than the measured latency threshold.

\section{Experiments}
\subsection{Experiment Setting}

\noindent\textbf{Evaluation settings: }We evaluate under two configurations:  1) online setting, where the perception models operate in real time, incorporating live sensor data at 10HZ, with communication latency and asynchronous message arrival to reflect realistic operational constraints, and 2) offline setting, which leverages static datasets with synchronized data and poses. The primary experiments are conducted in the online setting, while the offline setting is used to investigate the performance gap resulting from real-world challenges.

\noindent\textbf{Evaluation Metrics: } Following V2X-Real~\cite{xiang2024v2x}, we group different classes into three super-class (\ie, vehicle, pedestrian, and truck) as per their bounding box sizes. To assess the detection performance, we adopt mean Average Precision (AP) with Intersection-over-Union (IoU) thresholds of 0.3 and 0.5. Additionally, we confine the region of evaluation to [$-100$m, $100$m] along the x-axis and [$-40$m, $40$m] along the y-axis in the ego coordinate frame.

\noindent\textbf{Implementation Details: } PointPillar~\cite{lang2019pointpillars} is adopted as LiDAR backbones with voxel size of 0.4 meters in the x and y direction. For intermediate fusion methods, we select a compression rate of 32 to balance accuracy and bandwidth requirements, unless indicated otherwise. Following \cite{xiang2024v2x}, all the models are trained on the V2X-Real training split and evaluated on our proposed online dataset, which is derived from the test split of V2X-Real. We assess the online performance of these fusion strategies within a ROS~\cite{quigley2009ros} environment, thereby reflecting real-time operational constraints. For the online latency measurement and demonstration, each agent operates as an independent ROS node that ingests live sensor data, produces intermediate features, and communicates with collaborators via the Swarm ROS bridge~\cite{swarm}. For the perception performance benchmark, both the ego agent and collaborator’s ROS nodes run on the same computer, with the measured latency injected in online cooperative perception framework. This setup circumvents the synchronization inaccuracies and start-up  latency that emerge when simultaneously replaying ROS bags across physically separate systems. 


\noindent\textbf{Dataset modes: }We partition the online dataset into three distinct collaboration modes:  {V2X-RealLO-V2V}, where each vehicle receives transmitted information from other connected vehicles;  {V2X-RealLO-V2I}, where each vehicle obtains data from static infrastructure; and  {V2X-RealLO-I2I}, in which each infrastructure unit receives messages from other stationary infrastructure units.

\subsection{Benchmark methods}
\label{sec:benchmarked_methods}
We provide benchmarks for all three fusion strategies in cooperative perception, incorporating state-of-the-art (SOTA) methods as follows:

\begin{itemize}
    \item \textbf{No Fusion: }Each agent detects objects independently.
    \item \textbf{Late Fusion: }Detection outputs (\ie, bounding boxes and associated confidence score) are transmitted as standard \texttt{MarkerArray} ROS messages to neighboring collaborators. Upon receiving these outputs, the ego agent stores them in a memory buffer. After generating its own bounding box proposals, the ego agent retrieves the collected boxes and integrates them to produce consistent, fused predictions via NMS.
    \item \textbf{Early Fusion: }Each agent broadcasts its LiDAR point clouds as standard \texttt{PointCloud2} ROS messages, which the ego agent stores in a memory buffer. Upon receiving its own sensor data, the ego agent queries the buffer to retrieve collaborator point clouds, merging them into a holistic view for subsequent visual reasoning.
    \item \textbf{Intermediate Fusion: }Each agent broadcasts compressed intermediate features as ROS messages (Tab.~\ref{tab:ros_messages}). In this work, we benchmark three representative methods: F-Cooper\cite{chen2019cooper}, which employs max-pooling for feature fusion; AttFuse\cite{xu2022opv2v}, which applies self-attention for adaptive feature aggregation; and V2X-ViT~\cite{xu2022v2x}, which utilizes a multi-agent vision transformer for enhanced local and global context at a higher computational cost.

\end{itemize}

\begin{table*}[]
    \centering
    \begin{tabular}{c|c|cccccc|cc}
    \toprule
    \multirow{2}{*}{Models} &\multirow{2}{*}{Modes}& \multicolumn{2}{c}{AP$_{car}$@IoU} & \multicolumn{2}{c}{AP$_{ped.}$@IoU} &  \multicolumn{2}{c|}{AP$_{truck}$@IoU}& \multicolumn{2}{c}{mAP@IoU}  \\
     & & 0.3 & 0.5 & 0.3 & 0.5 & 0.3 & 0.5 & 0.3 & 0.5\\
     \midrule
        \multirow{2}{*}{Late Fusion}&Offline&47.4&44.4&29.2&14.9&18.7&9.1&31.8&22.8\\
        &Online&41.6\tiny\textcolor{BurntOrange}{-5.8}&32.0\tiny\textcolor{BurntOrange}{-12.4}&18.1\tiny\textcolor{BurntOrange}{-11.1}&8.4\tiny\textcolor{BurntOrange}{-6.5}&16.4\tiny\textcolor{BurntOrange}{-2.3}&9.4\tiny\textcolor{BurntOrange}{+0.3}&25.3\tiny\textcolor{BurntOrange}{-6.5}&16.6\tiny\textcolor{BurntOrange}{-6.2}\\
        \midrule
        \multirow{2}{*}{Early Fusion}&Offline&54.0&49.8&31.9&17.1&28.6&18.6&38.1&28.5\\
         &Online&41.0\tiny\textcolor{BurntOrange}{-13.0}&31.7\tiny\textcolor{BurntOrange}{-18.1}&17.4\tiny\textcolor{BurntOrange}{-14.5}&8.3\tiny\textcolor{BurntOrange}{-8.8}&20.3\tiny\textcolor{BurntOrange}{-8.3}&12.4\tiny\textcolor{BurntOrange}{-6.2}&26.2\tiny\textcolor{BurntOrange}{-11.9}&17.5\tiny\textcolor{BurntOrange}{-11.0}\\
         \midrule
           \multirow{2}{*}{F-Cooper~\cite{chen2019cooper}}&Offline&{58.8}&{51.4}&{27.0}&{12.1}&{39.2}&{32.6}&{41.7}&{32.1} \\
        &Online&{54.3}\tiny\textcolor{BurntOrange}{-4.5}&{43.0}\tiny\textcolor{BurntOrange}{-8.4}&{18.0}\tiny\textcolor{BurntOrange}{-9.0}&{6.7}\tiny\textcolor{BurntOrange}{-5.4}&{36.5}\tiny\textcolor{BurntOrange}{-2.7}&{27.0}\tiny\textcolor{BurntOrange}{-5.6}&{36.3}\tiny\textcolor{BurntOrange}{-5.4}&{25.6}\tiny\textcolor{BurntOrange}{-6.5} \\
        \midrule
        \multirow{2}{*}{AttFuse~\cite{xu2022opv2v}}&Offline&{58.1}&{55.2}&{33.1}&{16.9}&{35.1}&{28.5}&{42.1}&{33.5} \\
        &Online&{54.7}\tiny\textcolor{BurntOrange}{-3.4}&{45.9}\tiny\textcolor{BurntOrange}{-9.3}&{21.7}\tiny\textcolor{BurntOrange}{-11.4}&{9.6}\tiny\textcolor{BurntOrange}{-7.3}&{34.5}\tiny\textcolor{BurntOrange}{-0.6}&{26.6}\tiny\textcolor{BurntOrange}{-1.9}&{37.0}\tiny\textcolor{BurntOrange}{-5.1}&{27.4}\tiny\textcolor{BurntOrange}{-6.1} \\
        \midrule
        \multirow{2}{*}{V2X-ViT~\cite{xu2022opv2v}}&Offline&{57.8}&{56.0}&{36.4}&{19.3}&{48.2}&{43.9}&{47.4}&{39.8} \\
        &Online&{48.2}\tiny\textcolor{BurntOrange}{-9.6}&{40.3}\tiny\textcolor{BurntOrange}{-15.7}&{21.4}\tiny\textcolor{BurntOrange}{-15}&{9.1}\tiny\textcolor{BurntOrange}{-10.2}&{36.8}\tiny\textcolor{BurntOrange}{-11.4}&{30.5}\tiny\textcolor{BurntOrange}{-13.4}&{35.5}\tiny\textcolor{BurntOrange}{-11.9}&{26.6}\tiny\textcolor{BurntOrange}{-13.2} \\
        \midrule
        
    \end{tabular}
    \caption{\textbf{Performance comparisons between online and offline evaluations on V2X-ReaLO-V2V.} }
    \label{tab:offline_online}
\end{table*}

\subsection{Main Results}

\noindent\textbf{Communication Latency: } Table~\ref{tab:latency} summarizes the message sizes and transmission latencies for different fusion strategies. Notably, raw LiDAR point clouds, which comprise large volumes of unprocessed data, demand substantial bandwidth. In contrast, late fusion transmits only bounding box information, achieving the lowest latency. Intermediate fusion provides a middle ground by transmitting compact BEV feature maps, thus preserving richer information than late fusion without incurring the high latency associated with raw LiDAR data.

\noindent\textbf{Transmission module computation time: } To further evaluate the overhead introduced by the transmission module, Tab.~\ref{tab:transmission_latency} reports a breakdown of its key components. Overall, these additional costs remain moderate, sustaining real-time performance under typical Wi-Fi bandwidth and latency conditions. Notably, GPU–host memory transfers account for the largest portion of the delay—an aspect rarely considered in existing cooperative perception studies—while compression, serialization and packaging contribute comparatively minimal overhead.

\noindent\textbf{Inference speeds: }Tab.~\ref{tab:inference_speed} presents the average inference speed of our benchmarked cooperative methods. Most of the fusion approaches, including Late Fusion, Early Fusion, F-Cooper, and AttFuse, achieve inference times below 100ms, indicating their viability for real-time operations. By comparison, V2X-ViT exhibits a slightly higher inference time of 118.2ms, reflecting the additional computational complexity of its transformer-based architecture. This implies the difficulty of deploying  these computation-expensive intermediate fusion models in real-time and the need for further optimizations for these models to reach real-time performance. 

\noindent\textbf{Compression rate study: }In Tab.~\ref{tab:compression_rate}, we assess the influence of different compression ratios on message size and transmission latency for intermediate fusion. Higher compression ratios substantially decrease bandwidth demands, albeit at the potential cost of reduced feature fidelity. In practice, we find that a compression rate of 32× provides a reasonable balance, substantially lowering latency while preserving sufficient representational detail to support accurate cooperative perception.

\noindent\textbf{Benchmark results: }
Tab.~\ref{tab:benchmark} presents the benchmark results for three fusion strategies across V2V, V2I, and I2I collaboration modes. Overall, cooperative methods surpass the {No Fusion} baseline in V2I and I2I scenarios, while {Late Fusion} and {Early Fusion} tend to underperform in V2V. This discrepancy likely arises from both the ego vehicle and collaborator being in motion, which magnifies latency-induced misalignments and degrades the performance of classical Late and Early fusion methods. In contrast, the intermediate fusion methods {AttFuse} and {V2X-ViT} generally attain higher accuracy than other methods, whereas {F-Cooper} exhibits inferior results. We attribute this to F-Cooper’s simple max-pooling approach, which could be more vulnerable to feature misalignments under real-world asynchronous conditions. 
 Notably, intermediate fusion methods exhibit lower accuracy for detecting smaller objects (e.g., pedestrians) in V2V, suggesting that highly mobile targets are particularly susceptible to compounded spatial and temporal shifts. These findings highlight the promise of intermediate fusion for real-time collaboration, while also highlighting the importance of addressing real-world V2X constraints—particularly for smaller objects and highly mobile traffic participants—to fully harness robust V2X collaboration in real-world settings.
\begin{figure}[t]
\centering
    \includegraphics[width=0.7\linewidth]{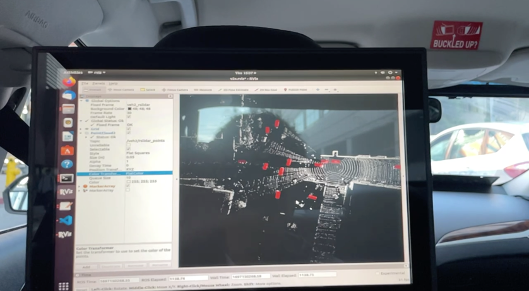}
    \caption{\textbf{Qualitative results for Intermediate Fusion}, deployed and running in real time on an in-vehicle computing platform in collaboration with a RSU. }
    \label{fig:onboard}
\end{figure}

\noindent\textbf{Offline and online performance comparisons: }In Tab.~\ref{tab:offline_online}, we compares performance under online and offline conditions for each cooperative fusion method. Overall, most approaches exhibit a noticeable drop in accuracy when transitioning to online deployment, reflecting the practical challenges posed by real-time constraints such as asynchronous message arrival, transmission latency and limited computation resources. This discrepancy is especially pronounced for small objects, underscoring their susceptibility to compounded temporal and spatial misalignments. Moreover, Intermediate Fusion methods AttFuse and F-Cooper tends to show slightly lower accuracy declines than Late Fusion and Early Fusion, demonstrating their robustness against real-world noises. Although V2X-ViT achieves the highest accuracy in offline settings, it undergoes the largest performance decline online. We attribute this to its higher computational overhead (see Tab.~\ref{tab:inference_speed}), which can amplify spatial and temporal discrepancies when compounded by real-world constraints such as latency and pose estimation errors. This result underscores the importance of jointly optimizing a cooperative perception model’s visual reasoning capabilities and computational efficiency to better accommodate real-time deployment challenges.

\section{Qualitative results}
Fig~\ref{fig:onboard} shows 3D bounding box detections generated by the AttFuse method in a real-world setting. On the vehicle side, an onboard computing platform processes LiDAR data, while a standard PC connected to the roadside sensor suite manages infrastructure-side computations. These distributed agents communicate via wireless Wi-Fi network, enabling timely data exchange under practical bandwidth and latency constraints.  The LiDAR sensor streams at 10 Hz, and we observe that the final detection outputs are also produced at 10 Hz, demonstrate the feasibility of operating Intermediate Fusion methods in real-world urban environments. More visualizations can be found in supplementary.


\section{Conclusion}
In this work, we present V2X-ReaLO, a novel open online V2X cooperative perception framework and  benchmark dataset for real-time V2X evaluation.  The framework unifies Early, Late, and Intermediate Fusion strategies within a single pipeline and is deployed on real vehicles and smart infrastructure, demonstrating the feasibility of V2X cooperative perception under real-world constraints. 
To extend beyond static datasets, we built an online benchmark dataset derived from V2X-Real, by incorporating additional ROS bags, utimatiely yeilding 25,028 frames, of which 6,850 are fully annotated key frames. V2X-ReaLO enables in-depth assessment of communication latency and overall perception performance under dynamic realistic conditions, which enables future online V2X research such as online cooperative perception for vulnerable road users. The associated dataset and benchmark codes will be released to facilitate future research on online cooperative perception.
{\small
\bibliographystyle{ieee_fullname}
\bibliography{egbib}
}

\end{document}